\documentclass[letterpaper]{article} 
\usepackage{aaai2027}  
\usepackage[hyphens]{url}  
\usepackage{graphicx} 
\usepackage{natbib}  
\usepackage{caption} 
\usepackage{booktabs}
\usepackage{multirow}
\usepackage{amsmath}
\usepackage{amssymb}

\providecommand{\modelname}{SkinCast}
\providecommand{\frameworkname}{SkinAdvisor}

\title{SkinAdvisor: An AI-Driven Decision Framework for Skin Production Planning via Multi-Modal Transformer Sales Forecasting}
\author{
    Ge Fan\textsuperscript{\rm 1},
    Kai Wang\textsuperscript{\rm 1},
    Chaoyun Zhang\textsuperscript{\rm 1},
    Yingjie Li\textsuperscript{\rm 2},
    Yuze Liu\textsuperscript{\rm 3},
    Junyang Chen\textsuperscript{\rm 4}\corresponding
}
\affiliations{
    \textsuperscript{\rm 1}Independent Researcher\\
    \textsuperscript{\rm 2}Tencent Inc.\\
    \textsuperscript{\rm 3}Swinburne University of Technology\\
    \textsuperscript{\rm 4}CSSE, Shenzhen University\\
    ge.fan@outlook.com, wangjinjie722@gmail.com, vyokky@163.com,\\
    wallaceyjli@tencent.com, yuzeliu@swin.edu.au, junyangchen@szu.edu.cn
}

\begin{document}

\maketitle

\begin{abstract}
Skins, cosmetic in-game items, are a primary revenue stream for modern
online games, yet production decisions (which skins to make, when to
release, how to price) remain driven by intuition and retrospective
analytics. We present \frameworkname{}, an AI-driven decision framework
for skin production planning, powered by \modelname{}, a deployed
multi-modal Transformer forecasting 180-day per-day sales trajectories
for unreleased skins. \modelname{} couples four encoders: a pretrained
Informer for player-preference time series, a pretrained
Player--Hero--Skin graph embedder for relational priors, and two
lightweight Transformers for skin attributes and release context, fused
through a shared backbone with a local-plus-global day-$x$ positional
encoding and decoded by an $N$-layer Informer in a single
non-autoregressive pass. \frameworkname{} plugs these forecasts into a
serving pipeline
(Select$\rightarrow$Fill$\rightarrow$Infer$\rightarrow$Sales$\rightarrow$ROI$\rightarrow$Ranking)
that ranks candidate skins by forecast-based ROI and supports
multi-candidate comparison. The system has been in production for a full
release cycle. Offline experiments show \modelname{} reduces RMSE by
$\sim\!20\%$ over the strongest baseline (Informer) and $\sim\!60\%$
over classical methods; stress tests on a 16-core CPU server confirm p99
latency under $8\mathrm{ms}$ and peak throughput above $1100$ rps. We
report two deployment lessons: order-of-magnitude accuracy suffices for
production decisions, and pretrained embeddings can leak temporal
information, requiring season-based splitting to prevent.
\end{abstract}

\section{Introduction}
\label{sec:intro}

\begin{figure}[t]
  \centering
  \includegraphics[width=0.495\linewidth,height=0.37\linewidth]{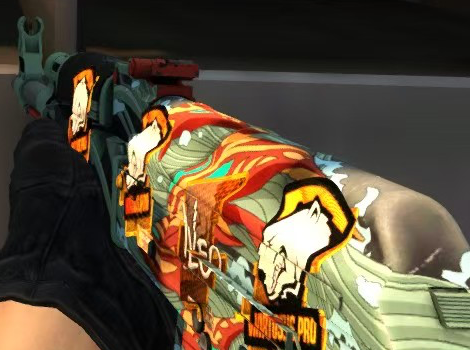}\hfill
  \includegraphics[width=0.495\linewidth,height=0.37\linewidth]{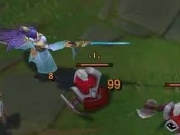}
  \caption{Examples of in-game skins, cropped from the authors' gameplay
  screenshots. Left: a weapon skin from \emph{CS:GO}. Right: a hero skin
  from \emph{League of Legends}.}
  \label{fig:teaser}
\end{figure}

Cosmetic skins are an important monetization mechanism in online games.
Before production begins, producers must choose the target hero, visual
tier, release channel, launch date, and production budget. These
decisions are difficult because the candidate skin has no historical
sales record and its demand depends on both player preferences and the
release context. As illustrated in Figure~\ref{fig:teaser}, skins take
different forms across genres, from lottery-style weapon finishes in the
FPS \emph{CS:GO} to directly purchased hero skins in the MOBA
\emph{League of Legends} (LoL), yet in both cases skin sales are a
primary revenue stream.

Producing a skin is resource-intensive: concept design, 3D modeling,
animation, VFX, sound, and QA all require substantial investment, and
sustained operations demand a steady cadence of new releases. In current
practice, decisions about \emph{which} skins to produce, \emph{when} to
release them, and \emph{how} to price them are driven largely by producer
intuition and retrospective analytics. This workflow fails to capture
rapidly shifting player preferences, underestimates the interaction
between release timing and preference trends, and often leads to missed
opportunities or over-production of low-return items.

Predictive modeling offers a data-driven alternative. Concretely, given
a candidate skin that has not yet been produced and all multi-source
signals available at a prediction time $t_p$ roughly one to two seasons
before its planned launch date $t_0$, the task is to forecast its daily
sales trajectory over the first 180 days after $t_0$. With such a
forecast in hand, producers can rank candidates by forecast-based return on
investment (ROI), allocate art and engineering capacity to the
highest-return concepts, and stress-test alternative release dates
against the seasonal demand curve. Deploying such a system in a live
studio, however, raises challenges that generic time-series or tabular
forecasters do not address:

\begin{itemize}
  \item \textbf{Heterogeneous, multi-source inputs.} Game-level time
  series (daily active users, match counts, hero usage), the
  player--hero--skin interaction graph over millions of entities, and
  sparse skin-attribute and release-context fields (series, tier, price,
  hero win rate, event tags) arrive in different formats and must be
  ingested jointly rather than flattened into one tabular snapshot.

  \item \textbf{Seasonal and calendar effects.} Skin sales show strong
  periodicity from weekly play cycles, seasonal holidays, and summer /
  winter breaks; a model that only sees ``days since launch''
  mis-estimates revenue at holiday peaks and off-season troughs,
  distorting the ROI ranking producers rely on.

  \item \textbf{Cold-start on unreleased skins.} The target item has no
  interaction history by construction~\cite{volkovs2017dropoutnet}, so
  the model must generalize from relational priors of its hero and
  attribute-level patterns of previously released skins.

  \item \textbf{Long-horizon forecasting at production scale.} Planning
  requires the full 180-day per-day trajectory, regenerated for thousands
  of candidate skins weekly under tight serving latency budgets.
\end{itemize}

We address these challenges with \modelname{}, a
multi-modal~\cite{baltrusaitis2019multimodal} Transformer forecaster,
together with \frameworkname{}, an ROI-driven decision framework built on
top of it. \modelname{} answers the input heterogeneity with four
specialized encoders: a pretrained Informer for macro time
series~\cite{zhou2021informer}, a pretrained Player--Hero--Skin graph
embedder for relational priors~\cite{fan2022pppne} that also anchors
cold-start generalization to the target skin's hero, and two lightweight
Transformer encoders for skin attributes and release context; a shared
backbone fuses them, and a local-plus-global day-$x$ positional encoding
carries the calendar signal into a single-pass non-autoregressive decoder
that emits the full 180-day trajectory in one forward call.
\frameworkname{} plugs these forecasts into a
Select$\rightarrow$Fill$\rightarrow$Infer$\rightarrow$Sales$\rightarrow$ROI$\rightarrow$Ranking
serving pipeline that ranks candidate skins by forecast-based ROI.
Figure~\ref{fig:arch} gives a joint overview: the top panel shows
\modelname{}'s encoder--decoder architecture, and the bottom panel shows
how \frameworkname{} wraps it into a producer-facing serving pipeline.

The system has been deployed in production and is used by producers to
prioritize the skin roadmap. In offline experiments, \modelname{} reduces
RMSE by $\sim\!20\%$ over the strongest baseline (Informer) and
$\sim\!60\%$ over classical methods; stress tests on a 16-core CPU server
show p99 latency under $8\mathrm{ms}$ at typical concurrency. We also
report two deployment lessons: the sufficiency of order-of-magnitude
accuracy for production decisions, and a temporal leakage via pretrained
embeddings that we diagnosed and fixed with season-based splitting.

Our contributions are:

\begin{itemize}
  \item We formulate skin sales forecasting as a multi-modal,
  long-horizon, per-day regression problem, and identify four concrete
  production challenges (heterogeneous inputs, calendar effects, cold
  start, and serving scale) that shape the design space.

  \item We propose \modelname{}, a deployed multi-modal Transformer
  forecaster that unifies a pretrained Informer time-series encoder, a
  pretrained Player--Hero--Skin graph encoder, and two lightweight
  Transformer encoders for static attributes and release context, with a
  local-plus-global day-$x$ positional encoding and a direct
  (non-autoregressive) Informer decoder for 180-day trajectories.
  Ablations identify the main contributing modules, with the
  time-series encoder being the most critical ($+22\%$ RMSE degradation
  when replaced by an equal-parameter MLP).

  \item We design and operate \frameworkname{}, an end-to-end ROI-driven
  decision framework that couples \modelname{} with a cost-allocation
  side path and a multi-candidate ranking service, turning it from a
  prediction tool into a production planning tool. The architecture could
  potentially be adapted to other content production scenarios (battle
  passes, seasonal events, fashion drops, collectible launches).

  \item We report offline accuracy, ablations, stress-test deployment
  metrics, and two concrete lessons learned from a full release cycle in
  a live game studio: (a)~order-of-magnitude correctness suffices for
  early-stage production decisions, and (b)~pretrained embeddings can
  leak temporal information indirectly, requiring season-based splitting
  to prevent.
\end{itemize}

\section{Data and Feature System}
\label{sec:feature}

\modelname{} consumes five feature families harvested from production
logs: four heterogeneous input families, one per encoder module of
Section~\ref{sec:encoder}, plus a target-day-indexed calendar family
that feeds the day-$x$ positional encoding of Section~\ref{sec:dayx}.
Below we state the two design constraints that shape the feature system;
the inputs to each module are described in Section~\ref{sec:encoder}.

\paragraph{Prediction-time freezing.} Producers make the production
decision one to two seasons, roughly 90 to 180 days, before a skin's
launch, so we define a \emph{prediction time} $t_p$ with
$t_0 - t_p \approx 90\text{--}180$ days (strictly $t_p < t_0$). Every
input, the macro time-series window, the graph snapshot, and the
release-context fields, is frozen as of $t_p$ so that no post-launch
information can leak into training or serving; the season-aligned
cutoff used to enforce this in training is described in
Section~\ref{sec:lessons}.

\paragraph{Family-to-module mapping.} The macro time-series family
feeds the pretrained Informer encoder (Module 1); the
player--hero--skin interaction graph feeds the pretrained PPPNE encoder
(Module 2) and is what resolves the cold-start challenge of
Section~\ref{sec:intro}; skin attributes feed the $\mathrm{TF1}$
encoder (Module 3); release context feeds the $\mathrm{TF2}$ encoder
(Module 4); and the calendar family is \emph{not} consumed by any
encoder but added to the pooled encoder summary as $\mathbf{p}_x$ before
the decoder stack.

\section{Proposed Method}
\label{sec:method}

\begin{figure*}[t]
  \centering
  \includegraphics[width=\textwidth]{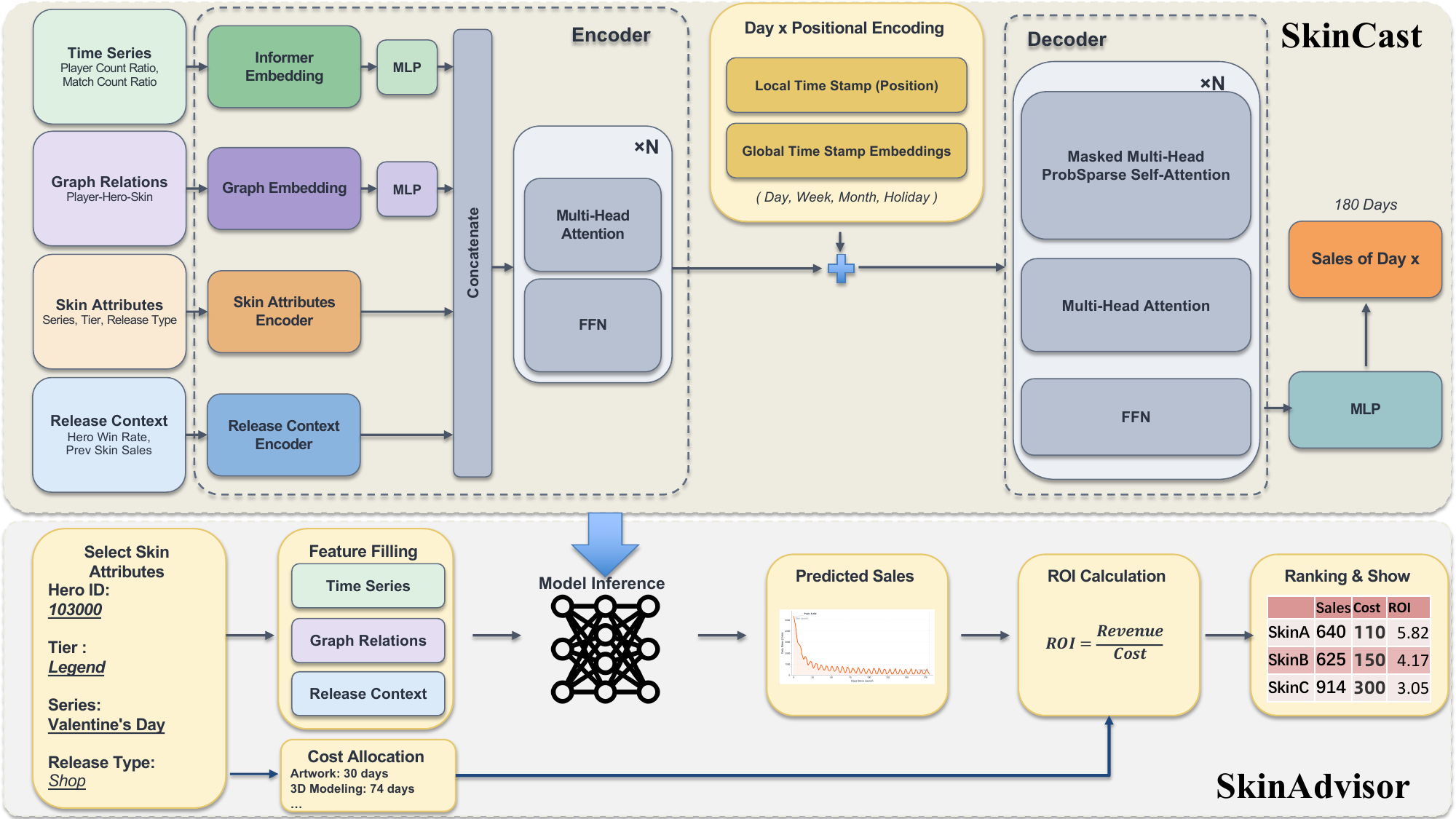}
  \caption{Overview of \modelname{} (top: multi-modal
  encoder--decoder architecture) and \frameworkname{} (bottom:
  producer-facing serving pipeline from core attributes to an
  ROI-ranked shortlist).}
  \label{fig:arch}
\end{figure*}

Given a candidate skin $s$ with release date $t_0$, \modelname{} predicts
the per-day sales trajectory
$\hat{\mathbf{y}}_{1:180} = (\hat{y}_1, \dots, \hat{y}_{180})$ over the
first 180 days after launch. The overall architecture
(Figure~\ref{fig:arch}) is a multi-modal encoder--decoder:

\begin{equation}
\mathbf{H}_{\mathrm{enc}}
\;=\;
\mathrm{Backbone}^{N}\!\big([\mathbf{H}_{\mathrm{ts}};\mathbf{H}_{\mathrm{gr}};
\mathbf{H}_{\mathrm{sa}};\mathbf{H}_{\mathrm{rc}}]\big),
\end{equation}
The concatenation above stacks four module vectors along the token
axis, so $\mathbf{H}_{\mathrm{enc}}\in\mathbb{R}^{4\times d}$.
We define a global summary
\begin{equation}
\mathbf g=\mathrm{Pool}_{\mathrm{fusion}}(\mathbf H_{\mathrm{enc}})
\in\mathbb R^d.
\label{eq:fusion_pool}
\end{equation}
The full encoder output serves as cross-attention memory; its pooled
summary initializes the daily decoder tokens. The head predicts
$\hat z_x$ on the log-sales scale, with $z_x=\log(1+y_x)$.
Let $\mathbf p_x\in\mathbb R^d$ be the target-day encoding defined in
Section~\ref{sec:dayx}. The decoder input is explicitly constructed as
\begin{equation}
\mathbf Z_0[x]=\mathbf g+\mathbf p_x,\quad x=1,\ldots,180,
\label{eq:decin}
\end{equation}
so $\mathbf Z_0\in\mathbb R^{180\times d}$. Throughout, $[x]$ denotes
the vector at sequence position $x$. The overall log-sales prediction is
\begin{equation}
\hat{\mathbf z}_{1:180}=\mathrm{MLP}_{\mathrm{head}}\!\left(
\mathrm{Dec}^{N}(\mathbf Z_0;\mathbf H_{\mathrm{enc}})\right).
\label{eq:overall}
\end{equation}
The head is applied independently at each target-day position;
the decoder's second argument supplies its cross-attention memory.
Here $\mathrm{Backbone}^{N}$ and $\mathrm{Dec}^{N}$ are stacks of $N$
Transformer encoder blocks~\cite{vaswani2017attention} and $N$ Informer decoder layers, and
$\mathbf{p}_x$ is the day-$x$ positional encoding defined below.

\subsection{Multi-Modal Encoder}
\label{sec:encoder}
Four specialized encoders map heterogeneous production inputs to a
shared $d$-dimensional space:
\begin{align}
\mathbf{H}_{\mathrm{ts}} &= \mathrm{MLP}_{\mathrm{ts}}\!\big(\mathrm{Pool}(\mathrm{InformerEnc}_{\theta^{*}}(\mathbf{X}_{\mathrm{ts}}))\big), \label{eq:ts}\\
\mathbf{H}_{\mathrm{gr}} &= \mathrm{MLP}_{\mathrm{gr}}\!\big(\mathrm{PPPNE}(\mathcal{G}_{\mathrm{PHS}})[h(s)]\big), \label{eq:gr}\\
\mathbf{H}_{\mathrm{sa}} &= \mathrm{Pool}\!\big(\mathrm{TF1}(\mathrm{Emb}(\mathbf{f}^{\mathrm{sa}}))\big), \label{eq:sa}\\
\mathbf{H}_{\mathrm{rc}} &= \mathrm{Pool}\!\big(\mathrm{TF2}(\mathrm{Emb}(\mathbf{f}^{\mathrm{rc}}))\big), \label{eq:rc}
\end{align}
where $\theta^{*}$ is a \emph{frozen} Informer encoder pretrained on
macro player-preference series, $\mathcal{G}_{\mathrm{PHS}}$ is the
Player--Hero--Skin interaction graph, $h(s)$ is the hero associated
with skin $s$, and $\mathrm{TF1}$ / $\mathrm{TF2}$ are two
parameter-separate instances of the same one-layer Transformer
architecture ($\mathrm{MHA}\!+\!\mathrm{FFN}$ with residual layer
normalization~\cite{ba2016layer}).

\paragraph{Time Series (Eq.~\ref{eq:ts}).} Encodes macro-temporal
signals that capture how player preferences toward heroes and skins
evolve, including DAU ratio, match-count ratio, hero-usage distribution,
and skin-usage frequency, over a sliding window of the $K$ days ending at
the prediction time $t_p < t_0$. Freezing $\theta^{*}$ and
training only $\mathrm{MLP}_{\mathrm{ts}}$ reuses a player-preference
embedder already maintained by the operations team, avoiding costly
joint pretraining on the small sales dataset.

\paragraph{Graph Relations (Eq.~\ref{eq:gr}).} Injects relational
priors mined from real player--hero--skin interactions via network
embedding~\cite{grover2016node2vec,he2020lightgcn,zhang2021gps}. Because the
embedding is pretrained on the full graph, a newly released skin
inherits signal from its hero's neighborhood; this is the primary
mechanism by which \modelname{} addresses the cold-start challenge of
Section~\ref{sec:intro}, even with zero direct behavioral trace of the
target skin, the hero embedding carries the size and purchasing
propensity of the fan base it will be exposed to.

\paragraph{Skin Attributes (Eq.~\ref{eq:sa}) and Release Context
(Eq.~\ref{eq:rc}).} Both static-feature encoders share the same
$\mathrm{Emb}\!+\!\mathrm{TF}\!+\!\mathrm{Pool}$ template (with $\mathrm{TF1}$ and $\mathrm{TF2}$ as separate parameter instances), treating each
categorical / numerical field as a token and letting self-attention
weigh cross-field interactions~\cite{fan2022fieldvae} (e.g.,
tier~$\times$~series, or price~$\times$~hero popularity). One Transformer
layer is sufficient because the field set is small
($F_{\mathrm{sa}}, F_{\mathrm{rc}} \!\sim\! 10$) and fully observed at
release time.

\paragraph{Backbone.} The four module tokens are concatenated and fed
through $N$ Transformer encoder blocks. Because they live in a shared
$d$-dimensional space, the backbone performs cross-modal attention
directly: the time-series token attends to the graph token, attribute
tokens attend to release-context tokens, and so on. This is where
higher-order interactions across heterogeneous sources are extracted,
in contrast to late-fusion baselines that only combine module-level
scalars.

\subsection{Day-$x$ Positional Encoding}
\label{sec:dayx}
\begin{equation}
\mathbf{p}_x
\;=\;
\underbrace{\mathbf{E}_{\mathrm{local}}(\mathrm{rel\_day}(x))}_{\text{local}}
\;\oplus\;
\underbrace{\textstyle\sum_{c \in \mathcal{C}} \mathbf{E}_{c}(c(x))}_{\text{global}},
\label{eq:pe}
\end{equation}
All embedding terms lie in $\mathbb R^d$, and $\oplus$ denotes
element-wise addition. Set $\tau_x=t_0+(x-1)\text{ days}$, with
$x=1$ denoting launch day. Here $\mathcal C=\{\text{day, wk, mo, hol}\}$
indexes day of month, day of week, month, and holiday status,
respectively, evaluated at $\tau_x$ using calendar information known
at prediction time.
The local term orders the horizon since launch and captures the
canonical revenue decay; the global term injects real-calendar
periodicity (weekly play cycles, seasonal holidays)
that a pure relative index cannot express. The encoding is added to
the pooled summary $\mathbf g$ once to initialize the decoder tokens.

\subsection{Informer Decoder and Training Objective}
\label{sec:decoder}
Starting from $\mathbf Z_0$ in Eq.~\ref{eq:decin}, the decoder produces
$\mathbf Z_N\in\mathbb R^{180\times d}$:
\begin{equation}
\mathbf{Z}_N = \mathrm{Dec}^{N}(\mathbf{Z}_{0};\mathbf H_{\mathrm{enc}}),
\qquad
\hat{z}_x = \mathrm{MLP}_{\mathrm{head}}(\mathbf{Z}_N[x]),
\label{eq:dec}
\end{equation}
\begin{equation}
\mathcal{L} \;=\; \frac{1}{180}\sum_{x=1}^{180}\big(\hat{z}_x - \log(1+y_x)\big)^{2}.
\label{eq:loss}
\end{equation}
Here $y_x\ge0$ is raw daily unit sales. Equation~\ref{eq:loss} is the
loss for one skin; training averages it over skins in each batch.
Equation~(\ref{eq:decin}) broadcasts the pooled encoder summary
$\mathbf g$ to every target day and adds that day's
positional embedding $\mathbf{p}_x$, yielding a 180-length decoder input
sequence $\mathbf{Z}_0$ in which each position carries both the global
multi-modal context and its own calendar / horizon position. Each
decoder layer then applies, in order, a masked ProbSparse self-attention
(causal over the 180 days), a cross-attention over
$\mathbf{H}_{\mathrm{enc}}$, and an FFN; the ProbSparse operator reduces
long-horizon cost from $\mathcal{O}(L^{2})$ to $\mathcal{O}(L \log L)$.

\paragraph{Direct (non-autoregressive) decoding.} \modelname{} emits all
180 days in a single decoder pass rather than feeding predictions back
as inputs. This choice was driven by an early autoregressive prototype
that accumulated error over the horizon and complicated serving.

\paragraph{Target transformation.} Raw daily sales are heavy-tailed.
We apply $y \!\mapsto\! \log(1\!+\!y)$ before Eq.~(\ref{eq:loss}).
Section~\ref{sec:exp} reports all metrics on this transformed scale;
serving converts predictions back to unit sales before aggregation:
\begin{equation}
\hat y_x=\max\{0,\exp(\hat z_x)-1\}.
\label{eq:inverse}
\end{equation}
These are point forecasts, not necessarily conditional mean sales.
Monotonicity preserves individual scalar orderings, but does not
guarantee cumulative-sales or ROI ranking quality.

\section{The \frameworkname{} Decision Framework}\label{sec:framework}

\frameworkname{} turns producer-specified skin attributes into a ranked
shortlist with 180-day sales curves and forecast-based ROI through the
pipeline in Figure~\ref{fig:arch}.

\subsection{End-to-End Pipeline}
\frameworkname{} executes six stages per producer request; the serving
half of Figure~\ref{fig:arch} (bottom panel) illustrates the full flow.

\paragraph{1. Select.} The producer opens the \frameworkname{} web
frontend and specifies the \emph{core} attributes of the candidate
skin: the associated hero, the product tier (e.g., legendary / epic /
rare), the release channel (in-game store, battle pass, seasonal event),
and the planned launch date. All four fields support multi-select so
that the producer can pose comparison queries such as ``same skin
concept across three candidate heroes'' or ``same hero across two
release channels''.

\paragraph{2. Fill.} The backend feature-completion service expands the
producer-specified core attributes into the full input bundle consumed
by \modelname{}. Skin series, theme, and effects flag are inferred from
a rule table indexed by (tier, hero, channel) and maintained by the art
team; price is looked up in a pricing-policy table indexed by (tier,
channel); hero win rate is fetched from the live game-stats service;
previous-skin sales of the same hero come from the sales warehouse; the
IP-collaboration flag and event tag come from the marketing calendar;
the hero's node embedding is looked up in the representation store that
serves the pretrained PPPNE table; and the macro time-series window is
pulled from the operations dashboard. This stage is the bridge between
how a producer thinks (``a legendary summer-event skin for hero X'')
and what \modelname{} consumes (a fully-specified four-module input).

\paragraph{3. Infer.} The filled feature bundle is forwarded to the
deployed \modelname{} inference service, which emits the 180-day
per-day sales trajectory $\hat{\mathbf{y}}_{1:180}$. When the producer
selected $N$ candidate combinations in stage~1, the backend expands the
Cartesian product and batches all $N$ candidates into a single forward
pass (batch dimension $= N$) to amortize inference cost.

\paragraph{4. Sales.} The trajectory is aggregated to
$\mathrm{total\_sales} = \sum_{x=1}^{180} \hat{y}_x$. The system also
computes 30-, 90-, and 180-day subtotals so that producers can compare
short-term spikes against long-tail revenue.

\paragraph{5. CostAlloc $\rightarrow$ ROI.} In parallel with the
inference chain, a cost-allocation side path looks up the expected
person-days for the given tier in a rule table maintained by the
production-management team (e.g., legendary $= 120$ person-days, epic
$= 60$, rare $= 20$) and converts person-days to person-hours before multiplying by the studio's blended hourly rate
to obtain $\mathrm{Cost}$. ROI is then computed as
\begin{equation}
\mathrm{ROI} \;=\; \frac{\mathrm{total\_sales}\times\mathrm{unit\_price}}{\mathrm{Cost}}.
\label{eq:roi}
\end{equation}
Here Cost is strictly positive and unit\_price denotes revenue per unit
under a fixed-price assumption. Throughout this paper, ROI is defined
as predicted revenue divided by production cost.
The ROI formulation (Eq.~\ref{eq:roi}) decouples cost estimation from
inference: CostAlloc is deliberately a \emph{side path} that does not gate inference
and can be revised (e.g., when the studio updates its rate card) without
retraining or redeploying \modelname{}.

\paragraph{6. Ranking.} The $N$ candidate ROI values are sorted
descending, and the frontend renders the top-$K$ candidates as overlaid
180-day sales curves (color-coded by candidate) alongside a ranked table
showing hero, tier, channel, launch date, total\_sales, Cost, and ROI.
Producers can apply filters (e.g., ``only candidates with
$\mathrm{ROI} \!>\! 3$'') and export the shortlist to the roadmap
planning system.

\subsection{Multi-Candidate Comparison}
Multi-select produces $N=H\times T\times C\times L$ candidate
combinations of heroes, tiers, channels, and launch dates, capped at
$N_{\max}$ to bound serving cost. Each candidate is evaluated independently
within one batched forward pass, and the top-$K$ by ROI (default $K=20$)
are returned for comparison.

\subsection{Deployment Topology}
\modelname{} is served as a stateless CPU inference service behind a load
balancer; the feature-fill, cost-allocation, and ranking services are
CPU-only microservices; the frontend is a single-page web app. The
entire pipeline runs on CPU only, with no GPU required at either
training or serving time. Measured latency and throughput under
stress-test loads are reported in Section~\ref{sec:exp}.

\section{Experiments}\label{sec:exp}

Since the quality of \frameworkname{}'s ROI rankings depends directly
on forecast accuracy, we evaluate \modelname{} offline against seven
baselines, ablate its components, and stress-test the serving pipeline.

\subsection{Experimental Setup}

\paragraph{Dataset.} Our skin sales data comes from a live online game
with tens of millions of Daily Active Users (DAU), spanning over three
years of continuous operation. To prevent temporal leakage, we use the
three most recent skin \emph{releases} (each release batches multiple
skins launched on the same date; all have been live for more than 180
days) as the test set, together accounting for roughly 5\% of the total
skin corpus; all earlier releases form the training set. For each test
release, all model inputs are frozen as of its prediction time $t_p$
(Section~\ref{sec:feature}): the macro time-series window ends at $t_p$,
both pretrained components are loaded from historical versions available
at $t_p$ and trained only on seasons completed by $t_p$, and
historical Release Context statistics come from the latest such season.
Planned price and event fields reflect information known at $t_p$.
This matches the online serving condition,
where a producer queries the system one to two seasons before release.
The prediction target is the $\log(1 + y)$ transformed daily sales over
the first 180 days post-launch, consistent with the training objective in
Eq.~(\ref{eq:loss}).

\paragraph{Baselines.} Since the target skin has no sales history of its
own, all baselines operate in the same pre-release setting as
\modelname{}: they forecast from a \emph{proxy} signal constructed from
previously released skins, never from the target skin's own post-launch
sales. We compare against seven baselines covering classical statistical
methods, deep time-series models, and tree-based approaches.
\textbf{ARIMA}~\cite{box1970time} and \textbf{Prophet}~\cite{taylor2018prophet}
are each fitted on the aggregated historical sales sequence of
same-hero / same-series skins (Prophet additionally with holiday and
seasonality regressors) and extrapolated to the target's 180-day
horizon. \textbf{MLP}~\cite{rumelhart1986learning} operates over
concatenated flat features without sequence modeling but still emits
the full 180-day trajectory via a multi-output regression head;
\textbf{GBDT}~\cite{friedman2001greedy} consumes the same flat features
and fits one gradient-boosted regressor per target day to produce the
same 180-day trajectory; a two-layer \textbf{LSTM}~\cite{hochreiter1997long}
encodes macro time-series concatenated with static features and feeds a
dense head; \textbf{DeepAR}~\cite{salinas2020deepar} is an
autoregressive RNN forecaster trained globally across all skins; and
\textbf{Informer}~\cite{zhou2021informer} is the ProbSparse Transformer
forecaster trained end-to-end on the macro time-series features only
(without graph, skin attributes, or release context). All seven
baselines therefore output the same per-day 180-step sequence as
\modelname{} and are evaluated under identical metrics.

\paragraph{Metrics.} We report Mean Absolute Error (MAE) and Root Mean
Squared Error (RMSE) on the $\log(1 + y)$ scale, computed per day and
averaged over the full 180-day horizon. Lower is better for both. We
additionally report the relative improvement of \modelname{} over each
baseline: $\mathrm{Impr.} = (\mathrm{Baseline} - \mathrm{\modelname{}})
/ \mathrm{Baseline} \times 100\%$.

\subsection{Main Results}

Table~\ref{tab:main_results} reports the results.
\modelname{} attains the lowest RMSE and MAE among all compared models,
with a relative improvement of $\sim\!20\%$ over the strongest baseline
(Informer) and $\sim\!60\%$ over classical methods (ARIMA).
Deep sequence models (LSTM, DeepAR, Informer) achieve lower error than
flat models (MLP, GBDT) in this evaluation, indicating that temporal
dynamics in macro player-preference signals carry predictive value.
The gap between \modelname{} and the plain Informer baseline
isolates the benefit of multi-modal fusion: the Informer baseline
sees the same macro time series but lacks graph relations, skin
attributes, release context, and the local-plus-global day-$x$
positional encoding.
Classical statistical models (ARIMA, Prophet) perform worst because they
compress the same-hero / same-series history into a single aggregated
proxy series and extrapolate it, so they can model neither the multi-modal
interactions nor the per-day calendar effects that \modelname{} captures.
Given the limited test set (three release batches), these results are
best read as descriptive rather than statistically conclusive.
Because \frameworkname{} ultimately consumes the forecast as a
\emph{ranking} signal, we additionally evaluate per-day ranking quality:
on each target day the test SKUs are ranked by predicted same-day sales
and scored by NDCG@k, averaged over the 180 days;
\modelname{} attains the highest NDCG at every cutoff
($k = 1, 3, 5, 13$), confirming that its point-accuracy advantage
carries over to per-day sales ranking; cumulative revenue and ROI ranking require separate evaluation.

\begin{table}[t]
\caption{Offline comparison on the $\log(1+y)$ scale.}
\centering
\small
\begin{tabular}{lcccc}
\hline \hline
Model & RMSE & MAE & \shortstack{RMSE\\Impr.\ (\%)} & \shortstack{MAE\\Impr.\ (\%)} \\
\hline
ARIMA    & 0.5185 & 0.4075 & 60.15 & 60.29 \\
MLP      & 0.4594 & 0.3643 & 55.02 & 55.58 \\
GBDT     & 0.4430 & 0.3472 & 53.37 & 53.39 \\
Prophet  & 0.4315 & 0.3396 & 52.12 & 52.35 \\
LSTM     & 0.3495 & 0.2774 & 40.88 & 41.66 \\
DeepAR   & 0.2974 & 0.2331 & 30.53 & 30.59 \\
Informer & 0.2572 & 0.2018 & 19.67 & 19.79 \\
\hline
\modelname{} & \textbf{0.2066} & \textbf{0.1618} & --- & --- \\
\hline \hline
\end{tabular}
\label{tab:main_results}
\end{table}

\begin{figure}[t]
  \centering
  \includegraphics[width=\linewidth]{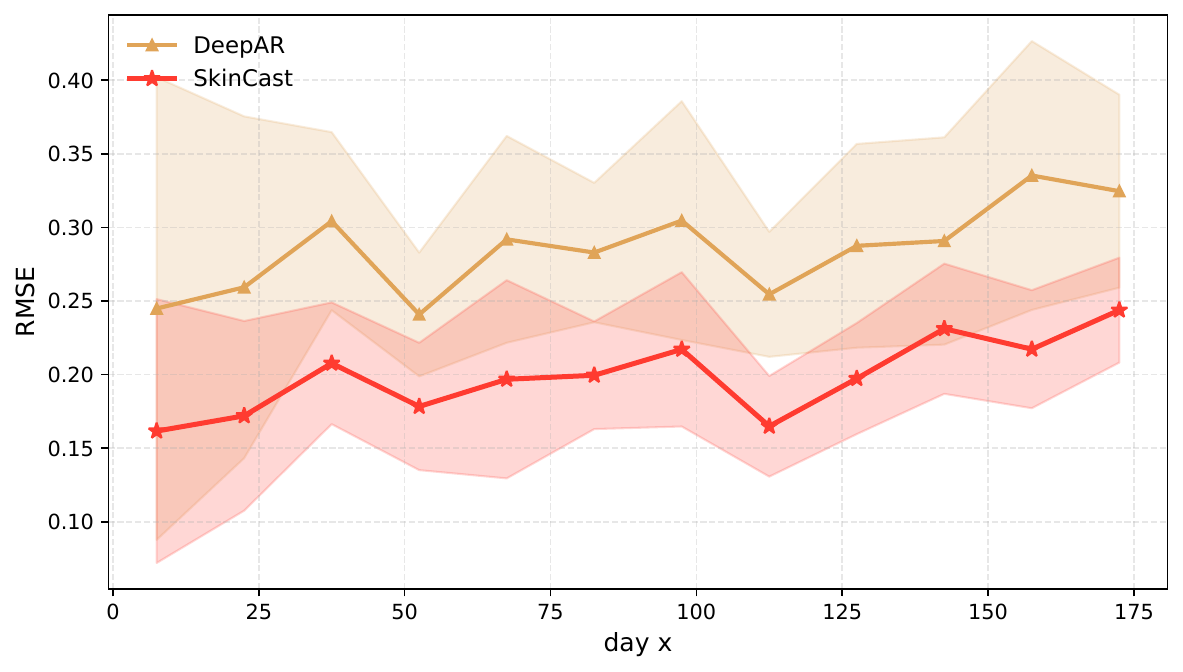}
  \caption{Per-day RMSE over the 180-day horizon (15-day bins) for
  \modelname{} and DeepAR; the shaded band shows the spread across test
  skins.}
  \label{fig:rmse_horizon}
\end{figure}

\paragraph{Horizon-wise error behavior.}
Table~\ref{tab:main_results} aggregates error over the full 180-day
horizon; Figure~\ref{fig:rmse_horizon} breaks it down by target day for
\modelname{} and the strongest autoregressive baseline (DeepAR). Two
patterns stand out. First, both models show lower RMSE in the first
weeks after launch and rising error toward the tail of the horizon. We
attribute this to the shape of skin demand: sales peak sharply in the
first days post-launch and decay into a long tail, so the regression
loss is dominated by the peak regime and the model fits the peak scale
more faithfully than the low-volume tail. Second, both curves exhibit
local oscillations that line up across the two models, which we
attribute to skin-specific operational events and calendar holidays
that perturb individual release trajectories. \modelname{} tracks these
perturbations with visibly smaller amplitude and a narrower spread
across test skins than DeepAR, consistent with its explicit calendar
encoding: the global (day-of-week / month / holiday) component of the
day-$x$ positional encoding lets \modelname{} absorb holiday- and
event-driven demand shifts rather than mistaking them for noise.

\subsection{Ablation Study}
We ablate one module at a time from the full \modelname{} and
re-evaluate on the same test set (Table~\ref{tab:ablation}). In every
``w/o X'' variant, the module X is replaced by a trainable MLP of
comparable parameter count while retaining its input signal.
This isolates the contribution of each module's \emph{architectural
choice} (pretrained Informer / pretrained PPPNE / Transformer encoder /
Transformer decoder / one-layer Transformer for static fields) rather
than the contribution of the raw input signal itself.

\begin{table}[t]
\caption{Ablation. ``w/o X'' replaces module X with a trainable MLP of
comparable parameter count.}
\centering
\footnotesize
\setlength{\tabcolsep}{4pt}
\begin{tabular}{lccc}
\hline \hline
Variant & RMSE & MAE & $\Delta$RMSE (\%) \\
\hline
\modelname{} (full)     & \textbf{0.2066} & \textbf{0.1618} & --- \\
\hline
w/o Time Series         & 0.2520 & 0.1987 & $+21.98$ \\
w/o Transformer Decoder & 0.2350 & 0.1867 & $+13.75$ \\
w/o Graph               & 0.2230 & 0.1727 & $+7.94$  \\
w/o Transformer Encoder & 0.2212 & 0.1749 & $+7.07$  \\
w/o TF2 (Rel.\ Context) & 0.2108 & 0.1691 & $+2.03$  \\
w/o TF1 (Skin Attr.)    & 0.2096 & 0.1685 & $+1.45$  \\
\hline \hline
\end{tabular}
\label{tab:ablation}
\end{table}

Replacing the pretrained Informer time-series encoder with an MLP
produced the largest RMSE increase ($+21.98\%$), followed by the
Transformer decoder ($+13.75\%$). Replacing the graph and backbone
encoder modules increased RMSE by $7.94\%$ and $7.07\%$, respectively.
The changes associated with the two static-feature encoders were much
smaller: replacing TF1 (Skin Attributes) and TF2 (Release Context) with
MLPs increased RMSE by only $1.45\%$ and $2.03\%$. This suggests that for
small, fully observed static field sets ($F_{\mathrm{sa}},
F_{\mathrm{rc}} \!\sim\! 10$), cross-field self-attention adds little
over an MLP, which is consistent with our choice of a single-layer
Transformer for these two modules; deeper variants were not evaluated.
The sequential, relational, and long-horizon-decoding pathways, by
contrast, benefit clearly from Transformer-based components. Given the
small test set, these results should be interpreted as descriptive
rather than statistically conclusive.

\subsection{Deployment Metrics}
\frameworkname{} serves an internal team of product planners; peak
concurrency is therefore low relative to consumer-facing services. We
stress-test the \modelname{} inference service to demonstrate that
latency remains well within the producer-facing UI budget
(p95 $< 3\mathrm{s}$) even under artificially high load.

\begin{figure}[t]
  \centering
  \includegraphics[width=\linewidth]{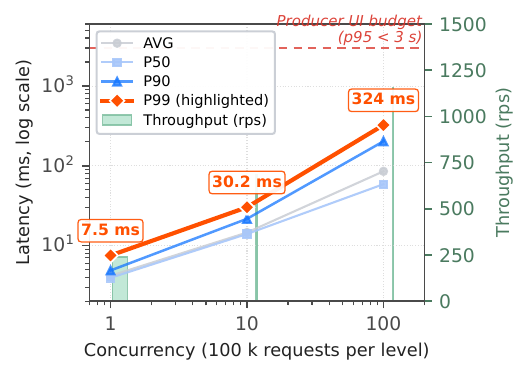}
  \caption{Inference latency (left axis) and throughput (right axis)
  versus concurrency on a single 16-core CPU server, 100k requests per
  level.}
  \label{fig:latency}
\end{figure}

Figure~\ref{fig:latency} shows that at concurrency~1 (the typical
producer workload), model inference completes in $\sim\!4\mathrm{ms}$
per request with p99 $< 8\mathrm{ms}$. Even at concurrency~100, far
beyond realistic internal usage, p99 remains under
$325\mathrm{ms}$, leaving ample headroom for the feature-fill,
cost-allocation, and ranking stages in the full pipeline. Peak throughput
exceeds $1100$ requests/second at concurrency~100, confirming that a
single 16-core CPU server comfortably supports the entire producer team
without queuing. The entire inference pipeline runs on CPU only, with no GPU required at
serving time, because the model is compact (the Informer encoder
operates over a short $K$-day window, and the decoder generates 180
tokens in a single pass) and the PPPNE embeddings are precomputed and
cached. This eliminates GPU provisioning costs and simplifies
operations, making the system practical for studios without dedicated
ML infrastructure.

\section{Lessons Learned}\label{sec:lessons}

Two lessons from the development and deployment of \frameworkname{}
shaped the final system and are likely applicable to other
forecasting-driven decision tools.

\subsection{``Good Enough'' Beats ``Precise''}
Through iterative discussions with producers and product planners, we
learned that \emph{order-of-magnitude correctness} matters far more than
point-estimate precision in the early stages of skin production
planning. Producers do not ask ``exactly how many units will this skin
sell on day~47?''; they ask qualitative questions such as ``if we sell
the same tier of skin for the same hero via a lottery system instead of
direct store purchase, how much does the expected revenue increase, one
order of magnitude or two?''

This insight reshaped our evaluation criteria and UI design. Rather than
optimizing solely for per-day MAE, we prioritize \emph{rank-order
consistency} between candidate skins (i.e., does the system correctly
identify that option~A outsells option~B by a factor of $\sim\!10$?).
The multi-candidate comparison mode of \frameworkname{}
(Section~\ref{sec:framework}) was a direct consequence: producers want
to see \emph{relative} curves side by side and make a directional call,
not to read a single number with a confidence interval. Formal ranking
metrics (Spearman correlation, pairwise accuracy) are planned for future
evaluation as more releases accumulate.

This experience highlights the importance of matching forecast precision
to the granularity of the production decision.

\subsection{Temporal Leakage via Pretrained Embeddings}
In early versions of \modelname{}, we observed suspiciously low training
loss and strong in-sample accuracy, yet poor generalization to the held-out
test skins. Root-cause analysis revealed a subtle form of
\textbf{temporal leakage through the pretrained Informer encoder}.

\paragraph{The problem.} The Informer time-series encoder was pretrained
on macro player-preference data that extended \emph{beyond} each skin's launch
date. From a data-construction perspective there was no direct leakage:
the pretraining target was next-step player-preference metrics, not skin
sales. However, the pretrained encoder had implicitly absorbed
information about post-launch demand surges (because a popular skin
launch causes DAU spikes and match-count changes that the encoder saw
during pretraining). When this encoder was frozen and used as a feature
extractor inside \modelname{}, the downstream model could exploit these
leaked signals to ``predict'' sales of skins that the Informer had
already implicitly observed, producing overfitting without any explicit
label contamination.

\paragraph{The fix: season-based temporal splitting.} We leveraged domain
knowledge of the game's \emph{season} structure (each season lasts
$\sim\!2$--$3$ months) to impose strict temporal boundaries on
\emph{both} pretrained components. For each skin, let $s_p$ denote the
latest season completed by its prediction time $t_p$, rather than the
season immediately preceding launch:
\begin{itemize}
  \item The Informer encoder is pretrained on player-preference sequences
  from seasons $\le s_p\!-\!1$, with the pretraining target (label)
  drawn from season $s_p$; both inputs and labels are available by $t_p$.
  \item The PPPNE graph embedding is likewise trained only on
  player--hero--skin interaction data from seasons $\le s_p$.
  \item For each test release, evaluation loads the historical pretrained
  model and embedding versions available at $t_p$, with the above
  season cutoff; no later-trained versions are used, and no input or
  supervision target from after $t_p$ enters either pretraining pipeline.
\end{itemize}
Because the number of seasons is finite and well-defined, this scheme is
both easy to audit and efficient: each season provides a substantial
volume of time-series and graph data, so both pretrained components still
receive sufficient signal without ever ``seeing the future.''

\paragraph{Impact.} The season-based split removes the leakage path: no
pretrained component can observe interactions or metrics from after the
prediction time $t_p$, so the reported test-set numbers reflect a
genuine pre-release forecasting setting rather than an inflated
in-sample fit. We note this as a methodological correction rather than
reporting a before/after accuracy delta, since we did not retain a
cleanly matched leaky-vs-fixed comparison under identical settings; the
value of the fix is that it makes the evaluation protocol sound, not that
it delivers a specific quantified gain.

The broader lesson: when using pretrained models as frozen feature
extractors, one must audit not only the downstream train/test split but
also the pretraining corpus boundary. Indirect leakage through learned
representations is invisible to standard data-pipeline checks and
requires domain-specific temporal reasoning to detect.

\section{Related Work}\label{sec:related}

\subsection{Long-Horizon Forecasting with Transformers}
Transformer-based forecasters have become the dominant paradigm for
long-horizon time-series prediction, a task that also arises in
per-entity outcome modeling such as clinical survival
estimation~\cite{zhu2026deepossa}. Informer~\cite{zhou2021informer}
introduced ProbSparse attention to reduce complexity from
$\mathcal{O}(L^2)$ to $\mathcal{O}(L \log L)$; Autoformer~\cite{wu2021autoformer}
added decomposition blocks; FEDformer~\cite{zhou2022fedformer} moved to
frequency-domain attention; PatchTST~\cite{nie2023patchtst} showed
that patching input sequences into sub-series tokens further improves
long-range forecasting; and TFT~\cite{lim2021temporal} introduced
static-plus-dynamic feature gating for multi-horizon forecasting. On the non-Transformer side, DeepAR~\cite{salinas2020deepar}
trains a global autoregressive RNN across many related series,
N-BEATS~\cite{oreshkin2020nbeats} uses pure MLP stacks with doubly
residual connections, and Prophet~\cite{taylor2018prophet} fits additive
trend-plus-seasonality models with holiday regressors.

With the exception of TFT, which gates static and dynamic features
separately, the above operate primarily on univariate or multivariate
time-series inputs without relational graph structure or sparse
categorical item attributes; recent work has even questioned whether
Transformers are necessary at all for long-horizon forecasting, showing
that a plain linear model can match or beat them on several standard
benchmarks~\cite{zeng2023transformers, qi2025timehfbillionscaletimeseries}. None addresses the multi-modal
fusion problem that \modelname{} targets: jointly encoding macro
player-preference series, a heterogeneous interaction graph, sparse
item attributes, and release-context fields into a unified
representation for cold-start forecasting of items that do not yet
exist. Our work extends the Informer lineage by wrapping its encoder
and decoder inside a four-module multi-modal architecture with a
local-plus-global positional encoding, and by demonstrating that the
fusion yields $\sim\!20\%$ RMSE reduction over a plain Informer
baseline on the same data.

\subsection{AI-Driven Production Decision Support}
AI in games has been widely studied from the \emph{player} perspective:
skill assessment~\cite{fan2022quickskill}, matchmaking and position-aware team formation~\cite{fan2024cupid,wang2026champ}, player preference
prediction~\cite{fan2021predicting,fan2023multicrit,chen2023neural,chen2023topic}, match outcome
forecasting, and dynamic difficulty adjustment. On the
\emph{production} side (deciding which content to create, when to
release it, and how to price it), published systems are scarce, though
large-scale deployed recommendation and ranking systems at YouTube,
Google Play, Alibaba, and Tencent~\cite{covington2016deep, cheng2016wide, pi2020search, Choi2026Optimizing, zhang2026dream, shao2026ugr, fan2022mvhan, fan2026uniboost, fan2026waste}
demonstrate how ML models are operationalized at production scale in
industry. E-commerce demand forecasting for new products shares the
cold-start challenge, typically addressed via meta-learned user
preference estimators~\cite{lee2019melu} or item-attribute similarity,
but without the relational graph structure (player--hero--skin) that is
unique to games. Heterogeneous graph
Transformers~\cite{hu2020heterogeneous} offer a general architecture
for such multi-type relational data, though they are not tailored to
the cold-start forecasting problem. Retail assortment
optimization~\cite{ferreira2016analytics, Basavaraju2025DevelopingAD} combines forecasting with
inventory decisions but does not address the 180-day per-day trajectory
granularity that game producers require for ROI planning.

To our knowledge, \frameworkname{} is the first deployed system that
combines multi-modal sales forecasting with an end-to-end ROI ranking
pipeline for game content production planning. It differs from
player-facing game AI in that its end-users are internal producers
rather than players, and from e-commerce demand forecasting in that it
must handle a heterogeneous graph with cold-start items, a fixed
180-day horizon, and a multi-candidate comparison UX that turns
predictions into actionable production shortlists.

\section{Conclusion}\label{sec:conclusion}

We have presented \frameworkname{}, an AI-driven decision framework for
skin production planning in online games, powered by \modelname{}, a
multi-modal Transformer that forecasts 180-day per-day sales trajectories
for unreleased skins by jointly modeling temporal, relational,
attribute, and contextual signals. Offline experiments
demonstrate that \modelname{} reduces RMSE by $\sim\!20\%$ over the
strongest baseline (Informer) and $\sim\!60\%$ over classical methods,
with ablations identifying the main contributing modules.
The system has been deployed in production with stress-tested p99
latency under $8\mathrm{ms}$ at typical concurrency and over $1100$
requests/second at peak load.

The framework could extend to other content-production settings.
Future work will incorporate concept art, community sentiment, and
patch notes through vision-language encoders
\cite{Ayan2026Banerjee, Paul2026Automatic} and retrieval-augmented
knowledge to complement the current structured inputs.

\section*{Acknowledgements}
This work was supported by the Shenzhen Science and Technology Program
(JCYJ20240813141417023) and the Natural Science Foundation of Guangdong
Province of China (2025A1515010233).

\bibliography{aaai2027}

\end{document}